\documentclass[12pt]{amsart}
\usepackage{amsbsy,amssymb,amsmath,amsthm,amscd,amsfonts,latexsym,amstext,delarray,
amsmath,graphicx} 
\usepackage[margin=.8in]{geometry}
\usepackage{color}

\numberwithin{equation}{section}

\def\bE{{\mathbb E}}

\def\bK{{\mathbb K}}

\title{Persistent Topology of Syntax}
\author[A.Port, I.Gheorghita, D.Guth, J.M.Clark, C.Liang, S.Dasu, M.Marcolli]{Alexander Port, Iulia Gheorghita, Daniel Guth, John M.~Clark, Crystal Liang, \\ Shival Dasu, Matilde Marcolli}
\address{Division of Physics, Mathematics and Astronomy, 
Caltech, 1200 E. California Blvd. Pasadena, CA 91125, USA}
\email{aport@caltech.edu}
\email{igheorgh@caltech.edu}
\email{dguth@caltech.edu}
\email{jmclark@caltech.edu}
\email{clliang@caltech.edu}
\email{sdasu@caltech.edu}
\email{matilde@caltech.edu}

\date{}

\begin{document}
\maketitle

\begin{abstract}
We study the persistent homology of the data set of syntactic parameters
of the world languages. We show that, while homology generators behave
erratically over the whole data set, non-trivial persistent homology appears
when one restricts to specific language families. Different families exhibit different 
persistent homology. We focus on the cases of the Indo-European and
the Niger-Congo families, for which we compare persistent homology
over different cluster filtering values. We investigate the possible significance,
in historical linguistic terms, of the presence of persistent generators of the
first homology. In particular, we show that the persistent first homology generator we find
in the Indo-European family is not due (as one might guess) to the Anglo-Norman 
bridge in the Indo-European phylogenetic network, but is related to the
position of Ancient Greek and the Hellenic branch within the network.
\end{abstract}

\section{Introduction}

This project is part of a broader ongoing investigation into the use
of methods from data analysis to identify the presence of 
structures and relations between the syntactic
parameters of the world languages, considered either globally
across all languages, or within specific language families and
in comparative analysis between different families.

\smallskip

We analyze the SSWL database of syntactic structures of world languages, 
using methods from {\em topological data analysis}. After performing principal 
component analysis to reduce the dimensionality of the data set, we compute
the persistent homology. The generators behave erratically when computed
over the entire set of languages in the database. However, if restricted to
specific language families, non-trivial persistent homology appears, which
behaves differently for different families. We focus our analysis on the two
largest language families covered by the SSWL database: the Niger-Congo
family and the Indo-European family. We show that the Indo-European family
has a non-trivial persistent generator in the first homology. By performing
cluster analysis, we show that the four major language families in the database
(Indo-European, Niger-Congo, Austronesian, Afro-Asiatic) exhibit different
cluster structures in their syntactic parameters. 
This allows us to focus on specific cluster filtering values,
where other non-trivial persistent homology can be found, in both the Indo-European
and the Niger-Congo cases. 

\smallskip

This analysis shows that the Indo-European family has a non-trivial
persistent generator of the first homology, and two persistent generators
of the zeroth homology (persistent connected components), with substructures
emerging at specific cluster filtering values. The Niger-Congo family, on the
other hand, does not show presence of persistent first homology, and
has one persistent connected component. 

\smallskip

We discuss the possible linguistic significance of persistent connected
components and persistent generators of the first homology. We propose
an interpretation of persistent components in terms of subfamilies, and
we analyze different possible historical linguistic mechanisms that may give
rise to non-trivial persistent first homology. 

\smallskip

We focus on the non-trivial persistent first homology generator in the
Indo-European family and we try to trace its origin in the structure 
of the phylogenetic network of Indo-European languages. The first
hypothesis we consider is the possibility that the non-trivial loop in the 
space of syntactic parameters may be a reflection of the presence 
of a non-trivial loop in the phylogenetic network, due to the historical
``Anglo-Norman bridge" connecting French to Middle English, hence
creating a non-trivial loop between the Latin and the Germanic subtrees.
However, we show by analyzing the syntactic parameters of these
two subtrees alone that the persistent first homology is not coming
from this part of the Indo-European family. We show that 
it is also not coming from the Indo-Iranian branch. Moreover, we
show that adding or removing the Hellenic branch from the remaining
group of Indo-European languages causes a change in both
the persistent first homology and the number of persistent component.

\medskip
\subsection*{Acknowledgment} This work was performed within the activities of the last author's 
Mathematical and Computational Linguistics lab and CS101/Ma191 class at Caltech. The last author 
is partially supported by NSF grants DMS-1007207, DMS-1201512, and PHY-1205440.

\section{Syntactic parameters and Data Analysis}

The idea of codifying different syntactic structures through
{\em parameters} is central to the Principles and Parameters
model of syntax, \cite{Chomsky}, \cite{ChoLa}, within Generative Linguistics. 
In this approach,
one associates to a language a string of binary ($\pm$ or $0/1$ valued) variables, 
the syntactic parameters, that encode many features of its 
syntactic structures. Examples of such parameters include 
{\em Subject Verb}, which has the value $+$ when in a clause 
with intransitive verb the order Subject Verb can be used;
{\em Noun Possessor}, which has value $+$ when a possessor 
can follow the noun it modifies;  {\em Initial Polar Q Marker}, which
has value $+$ when a direct yes/no question is marked by a clause 
initial question-marker; etc.\footnote{See {\tt http://sswl.railsplayground.net/browse/properties}
for a list and description of all the syntactic parameters covered by the SSWL database.}
The ``Syntactic Structures of the World's Languages" (SSWL) database, which 
we used in this investigation, includes a set of $115$ different parameters,
(partially) mapped for a set of $252$ of the known world languages.

\begin{figure}
\begin{center}
\includegraphics[scale=0.62]{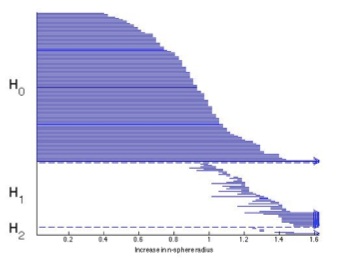}
\includegraphics[scale=0.62]{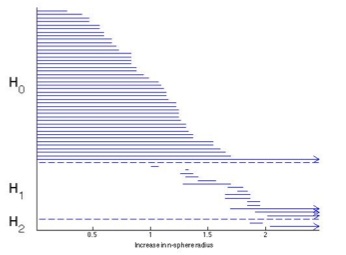}
\includegraphics[scale=0.55]{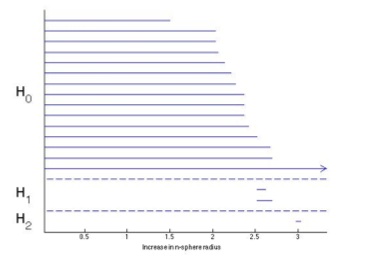}
\caption{Barcode graph over languages with $60\%$ 
of parameters known and $60\%$ of variance preserved; and with $80\%$ of
parameters known and $60\%$ of variance preserved. Third graph: barcode
for a random subset of 15 languages, $100\%$ of variance preserved.
\label{AllLing1}}
\end{center}
\end{figure}

\smallskip

The comparative study of syntactic structures across different world
languages plays an important role in Linguistics, see \cite{Sholpen} 
for a recent extensive treatment. In particular, in this study, we focus
on data of syntactic parameters for two of the major families of
world languages: the Indo-European family and the Niger-Congo
family. These are the two families that are best represented in the
SSWL database, which includes 79 Indo-European languages
and 49 Niger-Congo languages. The Niger-Congo family is the
largest language family in the world (by number of languages it
comprises). General studies of syntactic structures of Niger-Congo 
languages are available, see for instance \cite{BeSa}, \cite{MaRe}, 
though many of the languages within this family
are still not very well mapped when it comes to their syntactic parameters in the SSWL database.
The Indo-European family, on the other hand, is very extensively
studied, and more of the syntactic parameters are mapped. Despite
this difference, the data available in the SSWL database provide
enough material for a comparative data analysis between these
two families.

\smallskip

The point of view based on syntactic
parameters has also come to play a role in the study of Historical
Linguistics and language change, see for instance \cite{Galv}.
An excellent expository account of the parametric approach to syntax
is given in \cite{Baker}.

\smallskip

One of the sources of criticism to the Principles and Parameters model
is the lack of a good understanding of the space of syntactic
parameters, \cite{Hasp2}. In particular, the theory does not clearly identify a
set of independent binary variables that can be thought of as 
a ``universal set of parameters", and relations between syntactic
parameters are not sufficiently well understood. 

\smallskip

It is only in recent years, however, that accessible online databases of 
syntactic structures have become available, such as the WALS 
database of \cite{Hasp} or the SSWL database \cite{SSWL}.
The existence of databases that record syntactic parameters 
across different world languages for the first time makes them
accessible to techniques of modern {\em data analysis}. Our hope
is that a computational approach, based on various data analysis 
techniques applied to the 
syntactic parameters of world languages, may help the
investigation of possible dependence relations between 
different parameters and a better understanding of their overall structure. In the present
study, we focused on the data collected in the SSWL database,
and on {\em topological data analysis} based on {\em persistent homology}.

\smallskip

The structures we observe do not, at present, have a clear explanation
in terms of Linguistic theory and of the Principles and Parameters model
of syntax. The presence of persistent homology in the syntactic
parameter data, and its different behavior for different language
families begs for a better understanding of the formation and
persistence of topological structures from the Historical Linguistics
viewpoint, and from the viewpoint of Syntactic Theory. 

\begin{figure}
\begin{center}
\includegraphics[scale=0.41]{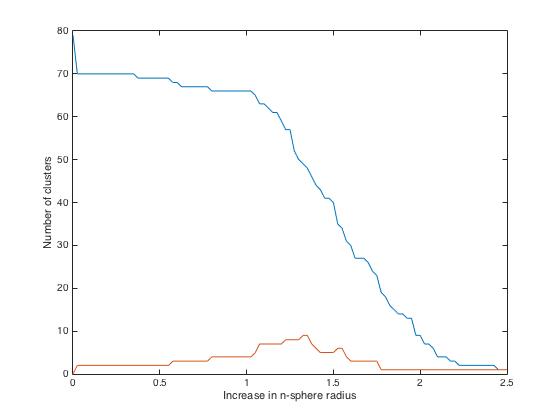}
\includegraphics[scale=0.41]{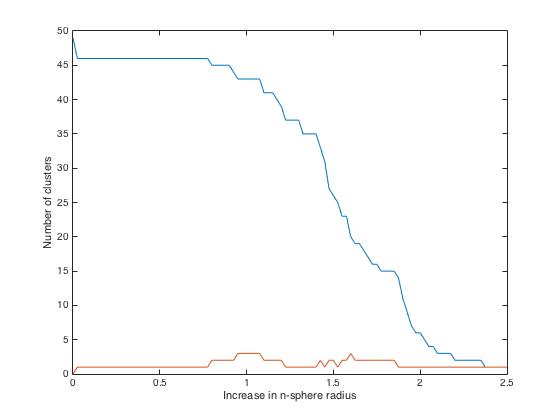}
\includegraphics[scale=0.41]{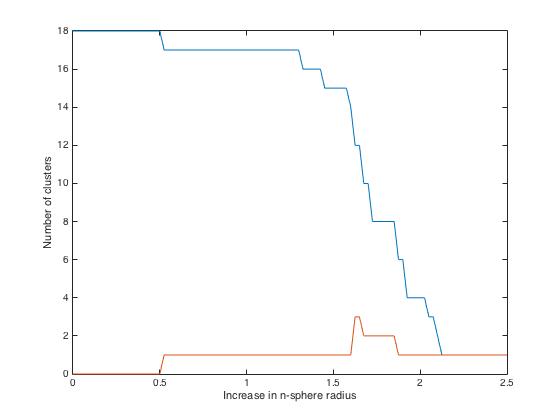}
\includegraphics[scale=0.41]{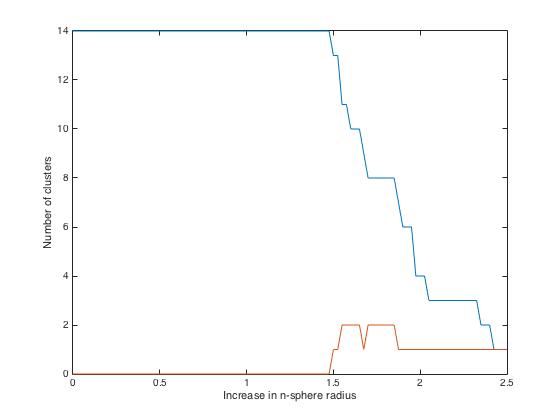}
\caption{Cluster structure of syntactic parameters for the Indo-European, the Niger-Congo, the Austronesian,
and the Afro-Asiatic language families. \label{ClusterFig}}
\end{center}
\end{figure}

\section{Persistent homology}

An important and fast developing area of data analysis, in recent years,
has been the study of high dimensional structures in large
sets of data points, via topological methods, see \cite{Carlsson},
\cite{EdHar}, \cite{Ghrist}. These methods of {\em topological
data analysis} allow one to infer global features from discrete 
subsets of data as well as find commonalities of discrete 
sub-objects from a given continuous object. The techniques
developed within this framework have found applications in
fields such as pure mathematics (geometric group theory,
analysis, coarse geometry), as well as in other sciences (biology,
computer science), where one has to deal with large sets of data. Topology
is very well-suited in tackling these problems, being qualitative in nature. 
Specifically, topological data analysis achieves its goal by transforming 
the data set under study into a family of simplicial complexes, indexed by a proximity parameter. One analyzes said complexes by computing their {\em persistent homology}, 
and then encoding the persistent homology of the data set in the form of a parametrized 
version of a Betti number called a {\em barcode graph}. Such graphs exhibit
explicitly the number of connected components and of higher-dimensional holes in the data. 
We refer the reader to \cite{Carlsson}, \cite{EdHar}, \cite{Ghrist} for a general overview
and a detailed treatment of topological data analysis and persistent homology.
As an example, persistent homology was used recently to study
the topology of a space of 3D images \cite{3D}, where the authors
determined that the barcode representation from persistent homology 
matched the homology of a Klein bottle.

\medskip
\subsection{The Vietoris-Rips complex}

Suppose given a set $X=\{ x_\alpha \}$ of points in some Euclidean
space $\bE^N$. Let $d(x,y)=\| x-y \|=(\sum_{j=1}^N (x_j-y_j)^2)^{1/2}$ 
denote the Euclidean distance function in $\bE^N$. 
The Vietoris-Rips complex $R(X,\epsilon)$ of scale $\epsilon$, over a field $\bK$, 
is defined as the chain complex whose space $R_n(X,\epsilon)$ of $n$-simplices corresponds 
to the $\bK$-vector space spanned by all the unordered
$(n+1)$-tuples of points $\{ x_{\alpha_0}, x_{\alpha_1}, \ldots, x_{\alpha_n} \}$
where each pair $x_{\alpha_i}, x_{\alpha_j}$ has distance
$d(x_{\alpha_i}, x_{\alpha_j})\leq  \epsilon$. The boundary maps 
$\partial_n: R_n(X,\epsilon) \to R_{n-1}(X,\epsilon)$, with $\partial_n \circ \partial_{n+1}=0$, are the
usual ones determined by the incidence relations of $(n+1)$ and $n$-dimensional
simplices. For $n\geq 0$, one denotes by 
$$ H_n(X,\epsilon) :=H_n(R(X,\epsilon),\partial) $$ $$ 
= {\rm Ker}\{ \partial_n: R_n(X,\epsilon) \to R_{n-1}(X,\epsilon)\} /
{\rm Range}\{ \partial_{n+1}: R_{n+1}(X,\epsilon) \to R_n(X,\epsilon)\} $$
the $n$-th homology with coefficients in $\bK$ of
the Vietoris-Rips complex. When the scale $\epsilon$ varies, one obtains
a system of inclusion maps between the Vietoris-Rips complexes,
$R(X,\epsilon_1) \hookrightarrow R(X,\epsilon_2)$, for $\epsilon_1 < \epsilon_2$.
By functoriality of homology, these maps induce corresponding morphisms
between the homologies, $H_n(X,\epsilon_1) \to H_n(X,\epsilon_2)$.
A homology class in $H_n(X,\epsilon_2)$ that is not in the image of $H_n(X,\epsilon_1)$
is a birth; a nontrivial homology class in $H_n(X,\epsilon_1)$ that maps to the zero element of
$H_n(X,\epsilon_2)$ is a death, and a nontrivial homology class in 
$H_n(X,\epsilon_1)$ that maps to a nontrivial homology class in $H_n(X,\epsilon_2)$
is said to persist. Mapping the deaths, births, and persistence of a set of generators
of the homology, as the radius $\epsilon$ grows gives rise to a barcode graph for the
Betti numbers of these homology groups. 
Those homology generators that survive only over short intervals of $\epsilon$ radii
are attributed to noise, while those that persist for longer intervals are considered to
represent actual structure in the data set.

\medskip
\subsection{Linguistic significance of persistent homology} 

When we analyze the persistent topology of different linguistic families 
(see the detailed discussion of results in \S \ref{TopLingSec}), we find
different behaviors, in the number of persistent generators in both $H_0$
and $H_1$. As typically happens in many data sets, the generators for
$H_n$ with $n\geq 2$ behave too erratically to identify any meaningful
structure beyond topological noise. 

\smallskip

In general, the rank of the $n$-th homology group $H_n$ of a 
complex counts the ``number of $(n+1)$-dimensional holes" that
cannot be filled by an $(n+1)$-dimensional patch. In the 
topological analysis of a point cloud data set, the presence of
a non-trivial generator of the $H_n$ at a given scale of the 
Vietoris-Rips complex implies the existence of a set of data
points that is well described by an $n$-dimensional set of
parameters, whose shape in the ambient space encloses
an $(n+1)$-dimensional hole, which is not filled by other 
data in the same set. In this sense, the presence of generators
of persistent homology reveal the presence of structure in the
data set. 

\smallskip

In our case, the database provides a data point for each recorded 
world language (or for each language within a given family), and
the data points live in the space of syntactic parameters, or in a
space of a more manageable lower dimension after performing 
principal component analysis. In this setting, the presence of
an ``$(n+1)$-dimensional hole" in the data (a generator of the
persistent $H_n$) shows that (part of) the data cluster around an
$n$-dimensional manifold that is not ``filled in" by other data points.
Possible coordinates on such $n$-dimensional structures represent
relations among the syntactic parameters, over certain linguistic
(sub)families. 

\smallskip

Since the only persistent generators we encountered
are in the $H_0$ and $H_1$, we discuss more in detail
their respective meanings.

\medskip
\subsection{Linguistic significance of persistent $H_0$}

The rank of the persistent $H_0$ counts the number of connected
components of the Vietoris-Rips complex. It reveals the presence of
clusters of data, within which data are closer to each other (in
the range of scales considered for the Vietoris-Rips complex) than to
any point in any other component. Thus, a language family exhibiting
more than one persistent generator of $H_0$ has linguistic parameters 
that naturally group together into different subfamilies. It is not known,
at this stage of the analysis, whether in such cases the subsets of
languages that belong to the same connected component correspond
to historical linguistic subfamilies or whether they cut across them.
We will give some evidence, in the case of the Indo-European family,
in favor of matching persistent generators of the $H_0$ to
major historical linguistic sub-families within the same family.
Certainly, in all cases, the connected components identified by different generators
of the persistent $H_0$ can be used to define a grouping into subfamilies, 
whose relation to historical linguistics remains to be investigated.

\medskip
\subsection{Linguistic significance of persistent $H_1$}

The presence of an $H_1$-generators also means that
part of the data (corresponding to one of the components
of the Vietoris-Rips complex) clusters around a one-dimensional
closed curve. More precisely, one can identify the
first homology group $H_1(X)$ of a space with the group
of homotopy classes $[X,S^1]$  of (basepoint preserving)
maps $f: X \to S^1$ to the circle. This means that, if there
is a non-trivial generator of the persistent $H_1$, then there
is a choice of a circle coordinate that best describes that part of
the data. The freedom to change the map up to homotopy
makes it possible to look for a smoothing of the circle coordinate. 
It is not obvious how to interpret these circles from the Linguistic
point of view. The fact that a generator of the $H_1$ represents
a $2$-dimensional hole means that, given the data that cluster along this
circle, no further data point determine a $2$-dimensional surface
interpolating across the circle. As the topological structures we
are investigating stem from a  Vietoris-Rips complex that measures
proximity between syntactic parameters of different languages,
we can propose a heuristic interpretation for the presence of such circles
as the case of a (sub)family of languages where each language
in the subfamily has other ``neighboring" languages with sufficiently 
similar syntactic parameters, so that one can go around the whole
subfamily via changes of syntactic parameters described by a single
circle coordinate, while parameter changes that move along 
two-dimensional manifolds and interpolate between data points on
the circle cannot be performed while remain within the same (sub)family. 

\begin{figure}
\begin{center}
\includegraphics[scale=0.6]{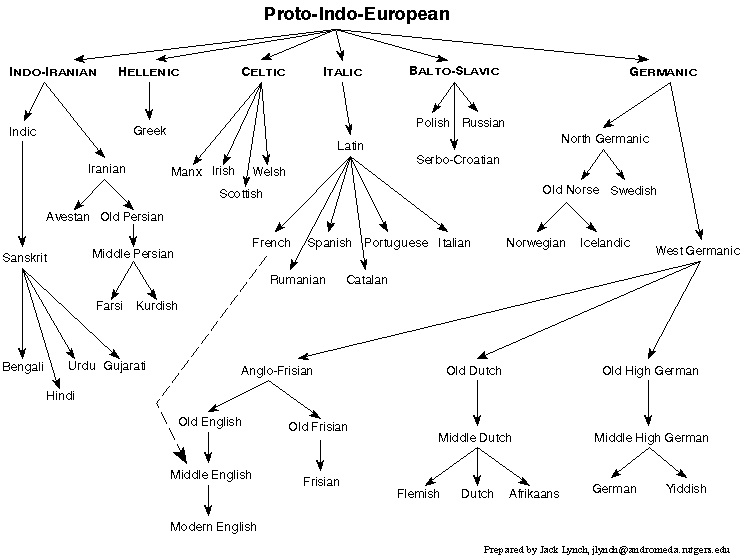}
\caption{The family ``tree" of the Indo-European languages (by Jack Lynch) showing the
loop between the Latin and the Germanic subtrees. \label{IEtreeFig}}
\end{center}
\end{figure}

Two different possible models of how a non-trivial generator of the persistent first
homology can arise point to different possible explanations in historical-linguistic terms.
\begin{figure}
\begin{center}
\includegraphics[scale=0.25]{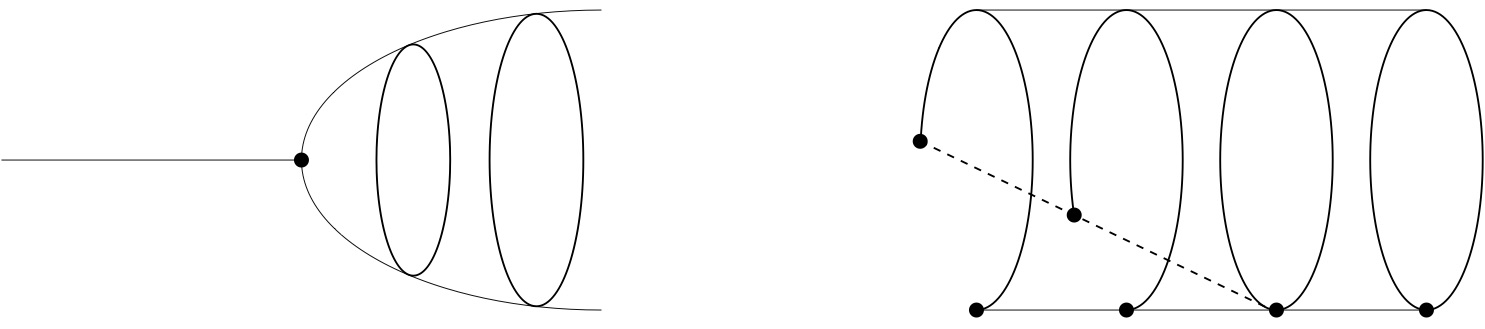} \ \ \ \ \ \ \ 
\caption{Two models of the development of a non-trivial loop in the space of parameters. \label{bifFig}}
\end{center}
\end{figure}
As shown in Figure \ref{bifFig}, the first model is a typical Hopf bifurcation 
picture, where a circle arises from a point
(with the horizontal direction as time coordinate). This model would be compatible
with a phylogenetic network of the corresponding language family that is a tree,
where one of the nodes generates a set of daughter nodes whose points in the
parameter space contain a nontrivial loop. The second possibility is of a
line closing up into a circle. This may arise in the case of a language
family whose phylogenetic network is not a tree, but it contains itself a loop that
closes off two previously distant branches. There are well known cases where
the phylogenetic network of a language family is not necessarily best described
by a tree. The most famous case is probably the Anglo-Norman bridge in the phylogenetic
``tree" of the Indo-European languages, see Figure \ref{IEtreeFig}. However, it is
important to point out that the presence of a loop in the phylogenetic network
of a language family does not imply that this loop will leave a trace in the syntactic
parameters, in the form of a non-trivial first persistent homology. Conversely, the
presence of persistent first homology, by itself, is no guarantee that loops may be 
present in the phylogenetic network, for example due
to possibilities like the Hopf bifurcation picture described above.
Thus, one cannot infer directly from the presence or absence of a persistent $H_1$
conclusions about the topology of the historical phylogenetic network.
The only conclusion of this sort that can be drawn is that a persistent $H_1$
suggests a phylogenetic loop as one of the possible causes. Conversely, one can 
read the absence of non-trivial persistent first homology as a suggestion
(but not an implication) of the fact that the phylogenetic network may be a tree
and that phenomena like the Anglo-Norman bridge did not occur
in the historical development of that family.

\smallskip

We will discuss this point more in detail in the case of the Indo-European
language family. This is a very good example, which shows how 
the possible correlation between loops in the space of
syntactic parameters and in the phylogenetic network is by no means an implication.
Indeed, the Indo-European language family contains both a known loop
in the phylogenetic network, due to the Anglo-Norman bridge (see Figure \ref{IEtreeFig}),
and a non-trivial generator of the persistent $H_1$. However, we will show using
our topological data analysis method that these two loops are in fact unrelated,
contrary to what intuition might have suggested. 

\begin{figure}
\begin{center}
\includegraphics[scale=0.41]{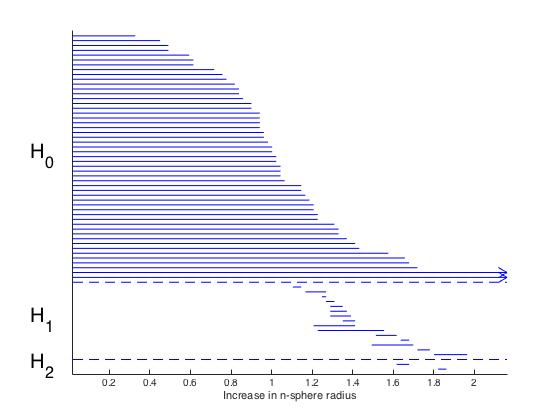}
\includegraphics[scale=0.41]{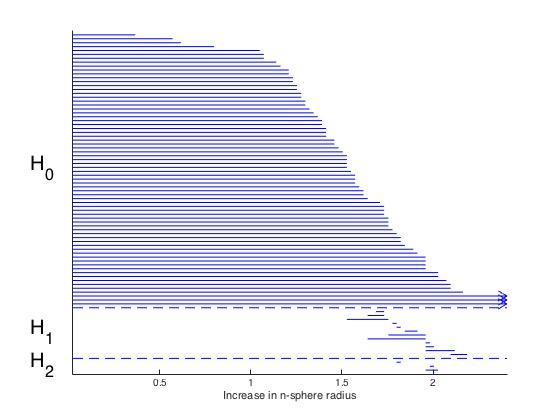}
\includegraphics[scale=0.6]{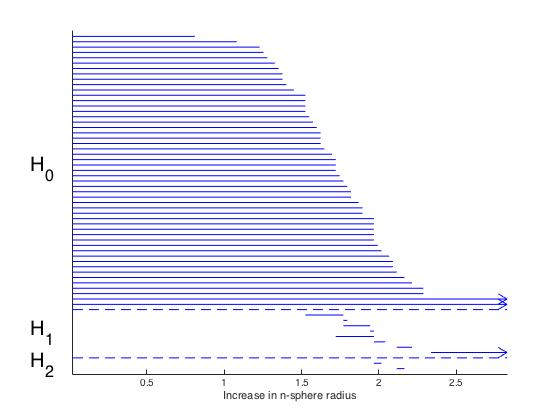}
\caption{Barcode diagram for the Indo-European language family, at indices $(7,5,96)$,
$(10,0,96)$, and $(10,5,98)$. \label{IE1fig}}
\end{center}
\end{figure}

\section{Data Analysis Procedure}\label{DataSec}

The SSWL  database \cite{SSWL} was first imported into a pivot table in Excel. 
The on-off parameters are represented 
in binary, in order to compute the distances between languages. 
However, the parameter values are not known for many of the languages in the
database: over one hundred of the languages have, at present, less than half of
their parameters known. Thus, we decided to replace empty language parameters 
with a value of 0.5. All together, we ended up with 252 languages, each 
with 115 different parameters.

\smallskip

We then proceeded to our analysis based on the results from Perseus
homology software \cite{Perseus}. This is achieved through a series of Matlab
scripts\footnote{A repository of the code used for this project is available at \newline
{\tt https://github.com/cosmicomic/cs101-project5}}. The script named {\tt data\_select\_full.m} allows for selection
of subsets of the raw data. It performs Principal Component Analysis
on the raw parameter data and saves it to a text file for use in Perseus.
The format of the data is that of a Vietoris-Rips complex. This script 
has two important parameters: a completeness threshold, and a percent
variance to preserve. The completeness threshold removes the languages
below a threshold of known parameters. The percent variance allows
us to reduce the dimensionality of our data.

\smallskip

The next script, named {\tt barcode.m}, was used to create barcode graphs
for data visualization. Perseus outputs the birth and death times
for each persistent homology generator, which are then used to construct
the barcode graph of the persistence intervals to visualize the structure
and determine the generators. The radii in our complexes are incremented 
by $1\%$ of the mean distance between languages.

\smallskip  

Data analysis was initially set up as a three step process: select the data
with the script {\tt data\_select\_full.m}, analyze it with Perseus, and use {\tt barcode.m}
to visualize the results. The final script, named {\tt run\_all}, streamlines
this process under a single input command.

\smallskip

Finally, our analysis includes examining how many data points belong to clusters
of points at any given radius. Clusters are constructed by creating
$n$-spheres of uniform radius centered at each data point. If the $n$-spheres
of two data points overlap, then those data points are in the same
cluster. A non-trivial cluster is a cluster with at least two data
points contained within. The scripts {\tt group\_select.m} and {\tt graph\_clusters.m}
make it possible to visualize the number of clusters and non-trivial clusters
as radius increases.

\smallskip

\begin{figure}
\begin{center}
\includegraphics[scale=0.41]{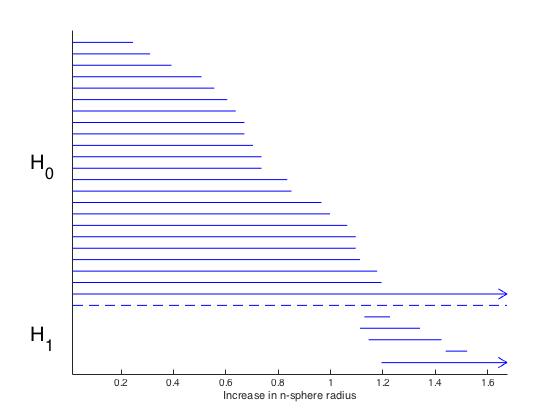}
\includegraphics[scale=0.41]{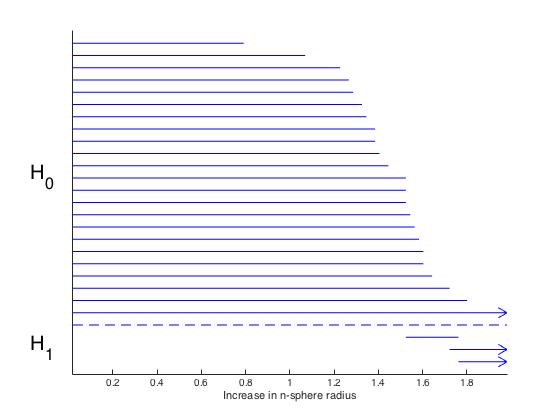}
\caption{Barcode diagram for the Indo-European language family, for 
cluster filtering value $165$ at indices $(7,5,95)$
and $(10,5,95)$. \label{IE3fig}}
\end{center}
\end{figure}

\section{The persistent topology of linguistic families}\label{TopLingSec}

A preliminary analysis performed over the entire set of languages in the SSWL
database shows that the non-trivial homology generators of $H_1$ and $H_2$ 
behave erratically. Moreover, there are too many generators of $H_2$ and $H_3$
to draw any meaningful conclusion about the structure of the underlying topological
space. One can see the typical behavior represented in Figure \ref{AllLing1}. In 
the first graph of Figure \ref{AllLing1} we included the languages with more than $60\%$ 
of the parameters known, while in the second we removed all languages with 
more than $20\%$ of the parameters unaccounted for. Here percentage of parameters is
with respect to the largest number of syntactic parameters considered in the SSWL
database. One can compare this with the case of a randomly generated subset of
languages, presented in the third graph of Figure \ref{AllLing1}. Notice that, while in the cases
represented in the first two graphs of 
Figure \ref{AllLing1} there is ``noise" in the $H_1$ and $H_2$ region,
that prevents a clear identification of persistent generators, the homology of random 
subsets of the data, as displayed in the third graph of Figure \ref{AllLing1}, 
is relatively sparse, 
containing only topologically trivial information. This observation lead us to the hypothesis
that the behavior seen in Figure \ref{AllLing1} stems from a superposition of some
more precise, but non-uniform, topological information associated to the various different
linguistic families. In order to test this hypothesis,
we decided to examine specific 
language families as an additional method of data filtering. We chose
the four largest families represented in the original database: Indo-European
with 79 languages, Niger-Congo with 49, Austronesian with 18, and
Afro-Asiatic with 14. Although some of the languages in the database
included latitude and longitude coordinates, these were ignored when
determining language family.

\begin{figure}
\begin{center}
\includegraphics[scale=0.41]{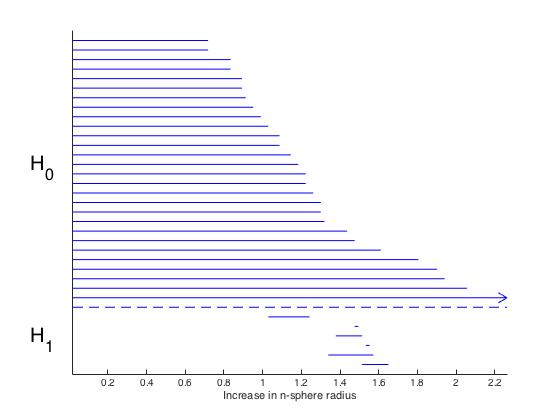}
\includegraphics[scale=0.41]{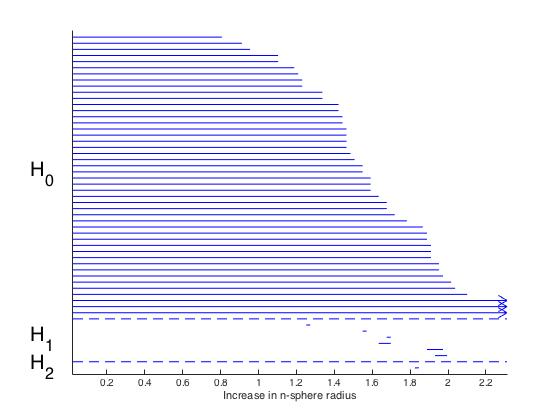}
\includegraphics[scale=0.6]{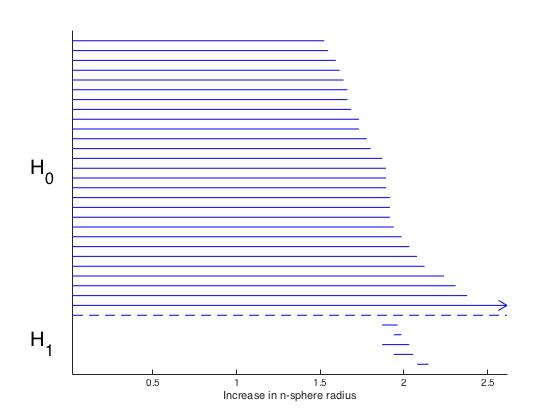}
\caption{Barcode diagram for the Niger-Congo language family, 
at indices $(7,5,107)$, $(10,0,100)$, and $(10,5,105)$. \label{NC1fig}}
\end{center}
\end{figure}

\medskip
\subsection{Cluster structures in major language families}\label{clustersec}

A first observation, when comparing syntactic parameters of different linguistic families, 
is that they exhibit different cluster structure of the syntactic parameters. 
This is illustrated in Figure \ref{ClusterFig},
in the case of the our largest families in the SSWL database. 

\smallskip

Based on this 
cluster analysis, we then focused on the cases of the Indo-European and
the Niger-Congo language family and we searched for nontrivial generators
of the first homology $H_1$ in appropriate ranges of cluster filtering. 

\smallskip

The cluster analysis of Figure~\ref{ClusterFig} suggests that
cluster filter values between $150$ and $200$ should provide 
interesting information. We computed additional barcode
diagrams corresponding to cluster filtering values $165$ and $190$.

\begin{figure}
\begin{center}
\includegraphics[scale=0.41]{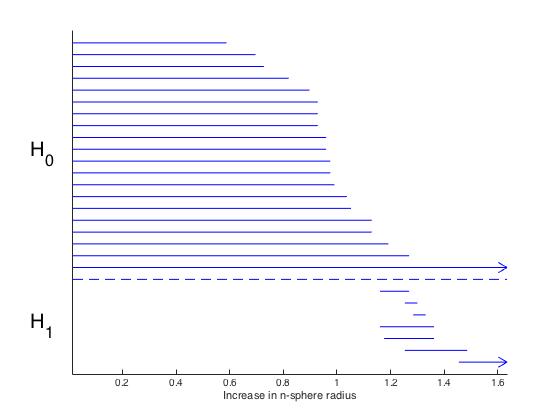}
\includegraphics[scale=0.41]{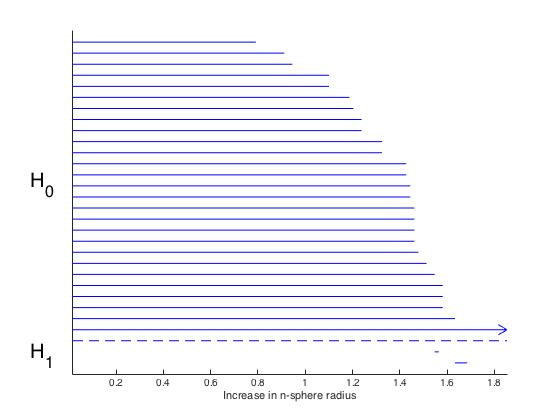}
\includegraphics[scale=0.41]{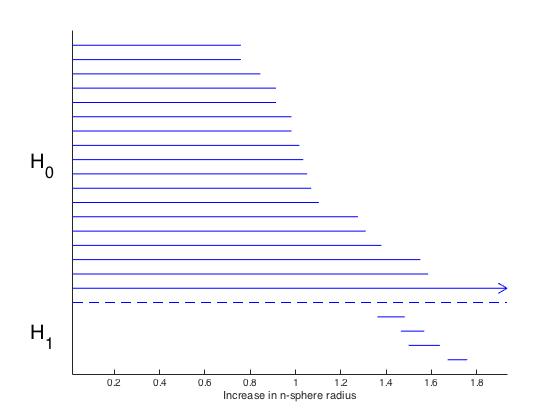}
\includegraphics[scale=0.41]{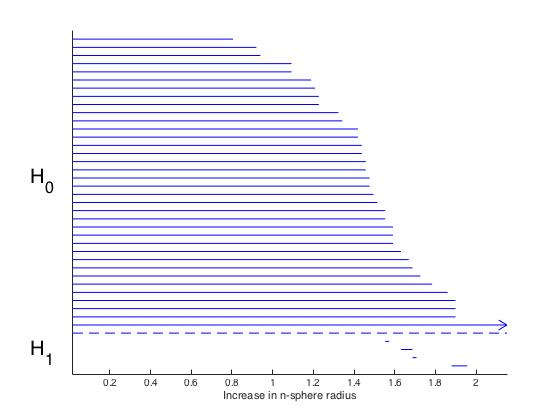}
\caption{Barcode diagram for the Niger-Congo language family, for 
cluster filtering value $165$ and 
indices $(7,3,100)$ and $(10,0,100)$, and for 
cluster filtering value $190$ and indices $(7,5,104)$ and
$(10,0,104)$. \label{NC2fig}}
\end{center}
\end{figure}

\smallskip
\subsection{Indexing in barcode graphs}
In the graphs presented in the following subsection, the barcode graphs
are labeled by a set of three indices. The first two indices refer to the
Principal Component Analysis and the third index to the runs of the Perseus program
computing births and deaths of homology generators of the Vietoris-Rips complex.
More precisely, the first index (7 or 10) refers
to the percent variance divided by 10, while the second index (0, 3 or 5)
refers to the percent complete divided by 10. They are discussed above in \S \ref{DataSec}.
The third parameter
is the number of steps in Perseus. If present, the additional parameter given by
the number after ``cluster" is one hundred times the radius used for cluster filtering.

\medskip
\subsection{Persistent topology of the Indo-European family}

We analyzed the persistent homology of the syntactic parameters for the
Indo-European language family. As shown in Figure \ref{IE1fig},
at values $(7,5,96)$ and $(10,0,96)$ one sees persistent generators of $H_0$
and intervals in the varying $n$-sphere radius, 
for which nontrivial $H_1$ generators exist. At values $(10,5,98)$, as shown in
Figure \ref{IE1fig}, one sees one persistent generator of $H_1$ and two persistent
generators of $H_0$. The existence of a persistent generator for the $H_1$ suggests
that there should be a ``circle coordinate" description for at part of the syntactic
parameters of the Indo-European languages. The fact that there are two persistent
generators of $H_0$ in the same diagram indicates two connected components, only
one of which is a circle: this component determines which subset of syntactic parameters 
admits a parameterization by values of a circle coordinate. 

\smallskip

Based on the cluster analysis described in \S \ref{clustersec} above, we then focused on 
specific regions of cluster filtering values that were more likely to exhibit interesting topology. 
For example, for cluster filtering value $165$, the results show, respectively, 
one generator of $H_0$ and one generator of $H_1$, for indices $(7,5,95)$, and
one generator of $H_0$ and a possibility of two persistent 
generators of the $H_1$, for indices $(10,5,95)$, see
Figure~\ref{IE3fig}. The appearance of persistent generators of the $H_1$ as
specific cluster filtering values identifies other groups of syntactic parameters that may
admit circle variable parameterizations. What these topological structures in the
space of syntactic parameters, and these subsets admitting circle variables description,  
mean in terms of linguistic theory remains to be fully understood. We analyze some
historical-linguistic hypotheses in the following subsection.

\medskip
\subsection{Indo-European persistent topology and historical linguistics}

It is often argued that the phylogenetic ``tree" of the family of Indo-European languages
should not really be a tree, because of the historical influence of French on Middle English,
see Figure~\ref{IE3fig}, which can be viewed as creating a bridge (sometimes referred to
as the Anglo-Norman bridge) connecting the Latin and the Germanic subtrees and
introducing non-trivial topology in the Indo-European phylogenetic network.
It is well known that the influx of French was extensive at the lexical
level, but it is not clear whether one should expect to see a trace of this historical phenomenon
when analyzing languages at the level of syntactic structures. It is, however, a natural question
to ask whether the non-trivial loop one sees in the persistent topology of syntactic
parameters of the  Indo-European family may perhaps be a syntactic remnant of the 
Anglo-Norman bridge.

\smallskip

\begin{figure}
\begin{center}
\includegraphics[scale=0.41]{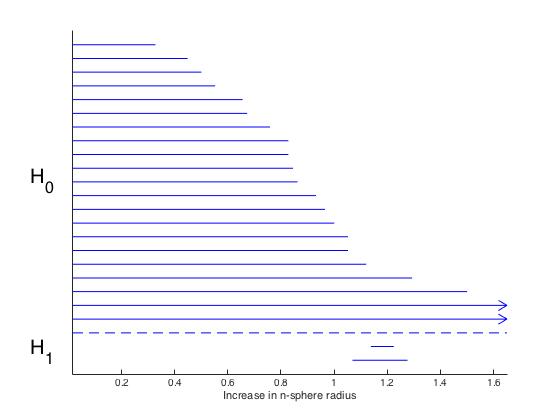}
\includegraphics[scale=0.41]{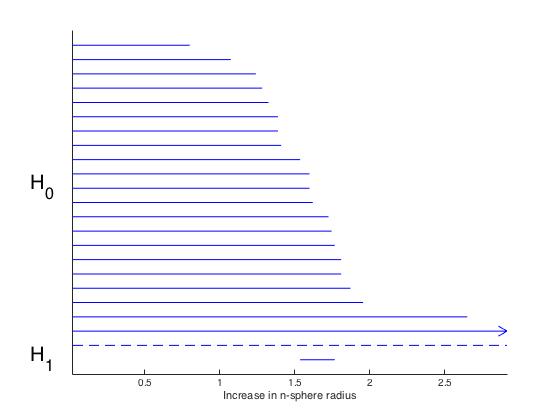}
\caption{Barcode diagram for the Latin+Germanic languages, 
at indices $(7,5,129)$ 
and $(10,5,130)$. \label{noANfig}}
\end{center}
\end{figure}

However, a further analysis of the SSWL dataset of syntactic parameters appears to
exclude this possibility. Indeed, we computed the persistent homology using only the
Indo-European languages in the Latin and Germanic groups. If the persistent generator
of $H_1$ were due to the Anglo-Norman bridge one would still find this non-trivial
generator when using only this group of languages, while what we find is that
the group of Latin and Germanic languages alone carry no non-trivial persistent first homology,
see Figure~\ref{noANfig}.

\smallskip

\begin{figure}
\begin{center}
\includegraphics[scale=0.6]{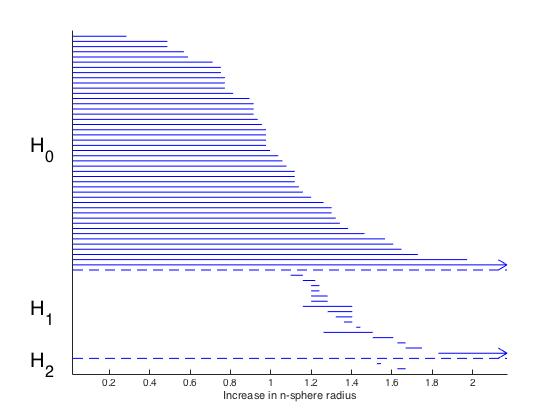}
\caption{Barcode diagram for the Indo-European family
with the Indo-Iranian subfamily removed, 
at indices $(7,5,97)$. \label{group3aFig}}
\end{center}
\end{figure}

In order to understand the nature of the two persistent generators of $H_0$,
we separated out the Indo-Iranian subfamily of the Indo-European family, to
test whether the two persistent connected components would be related to
the natural subdivision into the two main branches of the Indo-European family.
Even though the Indo-Iranian branch is the largest subfamily of Indo-European
languages, it is much less extensively mapped in SSWL 
than the rest of the Indo-European family, with only 9 languages recorded in the database.
Thus, a topological data analysis performed directly on the Indo-Iranian subfamily is less reliable,
but one can gain sufficient information by analyzing the remaining set of
Indo-European languages, after removing the Indo-Iranian subfamily. The result
is illustrated in Figure~\ref{group3aFig}. We see that indeed the number of
persistent connected component is now just one, which supports the proposal
of relating persistent generators of $H_0$ to major subdivisions into historical
linguistic subfamilies. Moreover, the persistent generator of the $H_1$ is still
present, which shows that the non-trivial first homology is not located in the 
Indo-Iranian subfamily.

\smallskip

\begin{figure}
\begin{center}
\includegraphics[scale=0.6]{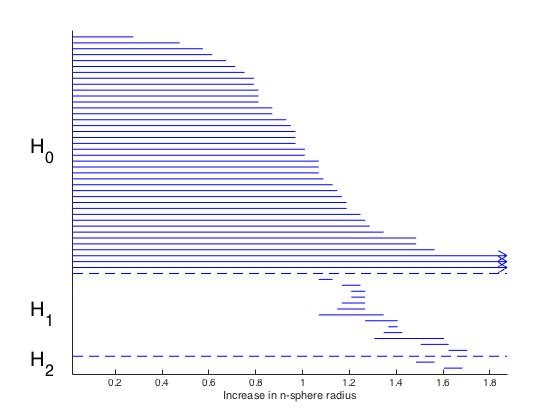}
\caption{Barcode diagram for the Indo-European family
with the Indo-Iranian and the Hellenic subfamilies removed, 
at indices $(7,5,96)$. \label{group3bFig}}
\end{center}
\end{figure}

In order to understand more precisely where the non-trivial persistent
first homology is located in the Indo-European family, we performed
the analysis again, after removing the Indo-Iranian languages and
also removing the Hellenic branch, including both Ancient and Modern
Greek. The resulting persistent topology is illustrated in Figure~\ref{group3bFig}.
By comparing Figures~\ref{group3aFig} and \ref{group3bFig} one sees that
the position of the Hellenic branch of the Indo-European family has a direct
role in determining the persistent topology. When this subfamily is removed,
the number of persistent connected components (generators of $H_0$) 
jumps from one to three, while the non-trivial single generator of $H_1$
disappears. Although this observation by itself does not provide an explanation
of the persistent topology in terms of historical linguistics of the Indo-European
family, it points to the fact that, if historical linguistic phenomena are involved
in determining the topology, they appear to be related to the role that 
Ancient Greek and the Hellenic branch played in the historical development of 
the Indo-European languages.

\smallskip

When performing a more detailed cluster analysis on the Indo-European
family, one finds sub-structures in the persistent topology. For instance, as
shown in Figure~\ref{IE3fig}, one sees a possible second generator of
the persistent $H_1$ for cluster filtering value 165, with indices $(10,5,95)$.
These substructures may also be possible traces of other historical linguistic
phenomena.

\medskip
\subsection{Persistent topology of the Niger-Congo family}

We performed the same type of analysis on the syntactic parameters of the 
Niger-Congo language family. The interesting result we observed is that the behavior of
persistent homology seems to be quite different for different language families.
Figure \ref{NC1fig} shows the barcode diagrams for persistent
homology at index values $(7,5,107)$, $(10,0,100)$, and $(10,5,105)$, which can
be compared with the diagrams of Figure \ref{IE1fig} for the
Indo-European family. In the Niger-Congo family, we now see persistent $H_0$ 
homology, respectively, of ranks $1$, $3$, and $1$ (compare with ranks $2$, $3$, $2$
in the Indo-European case). A lower rank in the $H_0$ means fewer connected
components in the Vietoris-Rips complex, which seems to indicate that the syntactic
parameters are more concentrated and homogeneously distributed 
across the linguistic family, and less ``spread out" into different sub-clusters. 


\smallskip

Following the cluster analysis of \S \ref{clustersec}, we also considered the
persistent homology for the Niger-Congo family at specific cluster filtering values.
While for cluster filtering value $165$ and indices $(7,3,100)$ one sees one
persistent generator of $H_0$ and a possibility of a persistent generator in the
$H_1$, cluster filtering value $165$ with indices  $(10,0,100)$, as well as
cluster filtering value $190$ with indices $(7,5,104)$ and
$(10,0,104)$ only show one persistent generator in the $H_0$.

\smallskip

This persistent homology viewpoint seems to suggest that syntactic
parameters within the Niger-Congo language family may be spread out more
evenly across the family than they are in the Indo-European case, with
a single persistent connected component, whereas the Indo-European 
ones have two different persistent connected component, one of which has
circle topology.

\section{Further questions}

We showed that methods from topological data analysis, in particular persistent
homology, can be used to analyze how syntactic parameters are distributed
over different language families. In particular we compared the cases of
Indo-European and Niger-Congo languages. 

\smallskip

We list here some questions
that naturally arise from this perspective, which we believe are worthy of
further investigation.
\begin{enumerate}
\item To what extent do persistent generators of the $H_0$ (that is,  the 
persistent connected components) of the data space
of syntactic parameter correspond to different (sub)families of languages
in the historical linguistic sense? For example, are the three $H_0$ generators
visible at scale $(10,0,100)$ in the Congo-Niger family a remnant of the historical 
subdivision into the Mande, Atlantic-Congo, and Kordofanian subfamilies? 
\item What is the meaning, in historical linguistic terms, of the circle components
(persistent generators of $H_1$) in the data space
of syntactic parameters of language families? 
Is there a historical-linguistic interpretation for the second $H_1$ generator
one sees at cluster filtering value 165 and scale $(10,5,95)$ in the Indo-European
family? Or for the $H_1$ generator one sees with the same cluster filtering, at scale
$(7,3,100)$ in the Niger-Congo case?
\item To what extent does persistent topology describe different
distribution of syntactic parameters across languages for different
linguistic families?
\end{enumerate}

\bigskip
\bigskip

\end{document}